\documentclass{article}

\usepackage[preprint]{neurips_2025}

\usepackage{xcolor}
\usepackage{color}
\definecolor{Gray}{gray}{0.95}
\definecolor{hookersgreen}{rgb}{0.0, 0.44, 0.0}
\definecolor{indiagreen}{rgb}{0.07, 0.53, 0.03}
\definecolor{islamicgreen}{rgb}{0.0, 0.56, 0.0}
\definecolor{kellygreen}{rgb}{0.3, 0.73, 0.09}
\definecolor{alizarin}{rgb}{0.82, 0.1, 0.26}
\usepackage{makecell}
\usepackage[normalem]{ulem}
\usepackage{pifont}
\newcommand{\cmark}{{\color{kellygreen} \ding{51}}}
\newcommand{\xmark}{{\color{alizarin} \ding{55}}}
\usepackage[utf8]{inputenc} 
\usepackage[T1]{fontenc}    
\usepackage{hyperref}       
\usepackage{url}            
\usepackage{booktabs}       
\usepackage{amsfonts}       
\usepackage{nicefrac}       
\usepackage{microtype}      
\usepackage{xcolor}         
\usepackage{xspace}
\usepackage{graphicx}
\usepackage{wrapfig}
\usepackage{amsmath}
\usepackage{multirow}
\usepackage{listings}
\usepackage{tcolorbox}
\usepackage{wrapfig}
\usepackage{subcaption}  
\definecolor{darkorange}{RGB}{255, 140, 0}
\definecolor{darkblue}{RGB}{84, 112, 198}
\definecolor{lightgreen}{RGB}{145, 204, 117}
\definecolor{lightyellow}{RGB}{250, 200, 88}
\definecolor{lightred}{RGB}{238, 102, 102}
\definecolor{lightblue}{RGB}{115, 192, 222}
\newtcolorbox{promptbox}[2][Prompt]{
colback=black!5!white,
arc=5pt, 
boxrule=0.5pt,
fonttitle=\bfseries,
title=#1, 
before upper={\normalsize}, fontupper=\fontfamily{ptm}\selectfont,
colframe=#2, 
}

\newcommand{\ie}{\emph{i.e.,}\xspace}

\newcommand{\eg}{\emph{e.g.,}\xspace}

\newcommand{\ignore}[1]{}
\usepackage{ifsym}

\def\dearena{\textsc{De-Arena}\xspace}

\title{Decentralized Arena: Towards Democratic and Scalable Automatic Evaluation of Language Models}

\author{%
  Yanbin Yin$^{1}$,~
  Kun Zhou$^{1}$\thanks{Corresponding authors.} ,~
  Zhen Wang$^{12*}$,
  Xiangdong Zhang$^{1}$,
  Yifei Shao$^{1}$,
  Shibo Hao$^{1}$,\\ \textbf{
  Yi Gu$^{1}$, 
  Jieyuan Liu$^{1}$,
  Somanshu Singla$^{1}$,
  Tianyang Liu$^{1}$,
  Eric P. Xing$^{2}$,
  Zhengzhong Liu$^{2}$,}\\
  \textbf{
  Haojian Jin$^{1}$,
  Zhiting Hu$^{1*}$
  }\\
  $^1$University of California, San Diego.~
  $^2$Mohamed bin Zayed University of AI.\\
  ~\texttt{\{kuzhou,zhw085,zhh019\}@ucsd.edu},\\
}

\begin{document}

\maketitle
\begin{abstract}
The recent explosion of large language models (LLMs), each with its own general or specialized strengths, makes scalable, reliable benchmarking more urgent than ever. Standard practices nowadays face fundamental trade-offs: closed-ended question-based benchmarks (\eg MMLU) struggle with saturation as newer models emerge, while crowd-sourced leaderboards (\eg Chatbot Arena) rely on costly and slow human judges.
Recently, automated methods (\eg LLM-as-a-judge) shed light on the scalability, but risk bias by relying on one or a few ``authority'' models. To tackle these issues, we propose Decentralized Arena (\dearena), a fully automated framework leveraging collective intelligence from all LLMs to evaluate each other. It mitigates single-model judge bias by democratic, pairwise evaluation, and remains efficient at scale through two key components: (1) a coarse-to-fine ranking algorithm for fast incremental insertion of new models with sub-quadratic complexity, and (2) an automatic question selection strategy for the construction of new evaluation dimensions.
Across extensive experiments across 66 LLMs, \dearena attains up to 97\% correlation with human judgements, while significantly reducing the cost. Our code and data will be publicly released on \url{https://github.com/maitrix-org/de-arena}.
\end{abstract}

\section{Introduction}

In recent years, the community has developed thousands of large language models (LLMs)~\cite{achiam2023gpt,touvron2023llama,bai2023qwen,liu2024deepseek} with ever-stronger general and specialized capabilities. To deploy these models in the real world effectively, we must assess and rank their performance accurately.
Concretely, existing work mostly collects a set of related high-quality questions, then judges the outputs of LLM to estimate the corresponding specialized capability~\cite{guha2024legalbench,xie2023pixiu,rajkumar2022evaluating}. By involving humans to vote on the preference of all LLM pairs (\ie deciding which LLM's output ``wins''), Chatbot Arena~\cite{chiang2024chatbot} provides robust and reliable leaderboards, yielding one of the most popular LLM benchmarks.

More importantly, with the increasing usage of LLMs in a variety of industrial applications and scientific tasks~\cite{hou2024large,taylor2022galactica,du2024large,thirunavukarasu2023large}, it’s crucial to evaluate LLMs’ capabilities on fine-grained dimensions, \eg math reasoning, physical sciences, and more specialized branches, such as algebra and astrophysics.
However, it is rather costly (both in terms of time and financially) for Chatbot Arena and other human-annotation-based benchmarks to support the evaluation of thousands of LLMs in thousands of fine-grained dimensions, \ie millions or even billions of human votes required. Moreover, although human judgments are still the gold standard, they can exhibit variability and subtle subjectivity, particularly when frontier models sometimes deploy persuasive ``sycophantic'' language~\cite{sharma2023towards} or other surface cues that may bias annotators toward incorrect but agreeable responses~\cite{schoch2020problem}.
To address them, researchers have also studied automatic evaluation methods, typically selecting one (or few) ``strongest'' LLM (\eg GPT-4) as a judge to evaluate all other model pairs~\cite{dubois2024length,li2023alpacaeval}.
However, the judge model can be biased, \eg by favoring outputs that resemble its own style~\cite{zheng2023judging, panickssery2024llm}. Optimizing models based on such evaluations could end up with overfitting to the biases of single judges.

\begin{wraptable}[19]{r}{0.55\textwidth}

  \caption{Comparison of representative LLM benchmarks based on the types of judge models, whether automatically evaluating LLMs, selecting the data, or using open-ended questions.}
  \centering
  \small
  \begin{tabular}{lcccc}
    \toprule
    \textbf{Benchmarks} & \textbf{Judges} & \makecell[c]{\textbf{Auto}\\\textbf{Eval}}  & \makecell[c]{\textbf{Auto} \\ \textbf{Data}} & \makecell[c]{\textbf{Open} \\ \textbf{ended}} \\
    \midrule
    \textbf{Compass Arena} & Human &  \xmark & \xmark & \cmark \\
    \textbf{Chatbot Arena} & Human &  \xmark & \xmark & \cmark \\
    \textbf{MixEval} & - & \cmark & \cmark & \xmark \\
    \textbf{LiveBench} & - & \cmark & \xmark & \xmark \\
    \textbf{Alpaca Eval} & GPT-4 & \cmark & \xmark & \cmark \\
    \textbf{WildBench} & GPT-4 & \cmark & \xmark & \cmark \\
    \textbf{BiGGen Bench} & GPT-4 & \cmark & \xmark & \cmark \\
    \textbf{PRD} & Five LLMs & \cmark & \xmark & \cmark \\
    \textbf{Auto Arena} & Five LLMs & \cmark & \cmark & \cmark \\
    \midrule
    \textbf{De-Arena} & All LLMs & \cmark & \cmark & \cmark \\
    \bottomrule
  \end{tabular}

  \label{tab:all_benchmark}
\end{wraptable}

To achieve the goal of reliable and scalable evaluation across various dimensions, we propose Decentralized Arena~(De-Arena), an automatic evaluation method based on the ``wisdom of the crowds''. Table~\ref{tab:all_benchmark} illustrates the main difference between De-Arena and other benchmarks. 
The core idea behind De-Arena is to use the collective intelligence of all LLMs to evaluate and compare themselves. 
This forms a democratic system where \emph{all LLMs to be evaluated are also judges to evaluate others}, leading to fairer rankings than the automatic methods relying on a few centralized ``authority'' judge models~\cite{salminen2015role,surowiecki2004wisdom}. This should address the single-judge bias or the bias from a similar model family~\cite{goel2025great}. 
Additionally, its automatic benchmarking process also supports \textit{scaling up the number of test LLMs and dimensions} with a lower cost than collecting large-scale human annotations.

To implement our De-Arena, a naive approach is to utilize all the LLMs to judge all other model pairs (similar to Chatbot Arena), based on manually crafted or selected high-quality questions.
However, it would lead to a prohibitively expensive time complexity of $\mathcal{O}(n^3 k)$\footnote{$n$ and $k$ are the number of models and questions, respectively.}. 
To make De-Arena a more efficient and fully automatic paradigm, we devised (1) \emph{the coarse-to-fine incremental ranking algorithm} and (2) \emph{the representative question selection strategy}.
Concretely, in our ranking strategy, the LLMs will be incrementally inserted into the rank list (\ie one by one), by first finding the rough position via binary search and then fine-grained in-window ranking.
Such a way naturally supports gradually growing the rank list by adding the latest LLMs, and the low complexity of binary search (\ie $\mathcal{O}(kn\log n)$) helps greatly reduce the time cost. We later empirically show that even the coarse-grained step (i.e., binary insertion) can achieve highly capable performance, thanks to the diverse set of judges. Besides efficiency, we introduce an adaptive weight mechanism and ELo system to re-weight judges dynamically (akin to PageRank), making De-Arena more reliable.

For question selection, our De-Arena leverages the above ranking strategy to select a few representative questions that lead to more consistent results, as the majority.
In this way, we ensure that the new evaluation dimensions can be automatically built by selecting a few high-value ones from the collected question candidates.
Based on the above designs in De-Arena, we automatically construct nine fine-grained dimensions, and efficiently evaluate 66 LLMs on them (as shown in Table~\ref{tab:bigtable}).
Experimental results demonstrate the effectiveness of our method, achieving up to 97\% correlation with human-annotation-based Chatbot Arena in the overall dimension (shown in Table~\ref{tab:eval-table}), with a cost similar to existing benchmarks (shown in Figure~\ref{fig:cost-study}).
Extensive studies also reveal that our method can significantly reduce the bias from a single-LLM judge (Table~\ref{tab:bias-study} and Table~\ref{tab:bias-rank}), and becomes more stable and accurate as the number of models increases (Figure~\ref{fig:convergence}), demonstrating its reliability and stability. Ablation study shows that the performance is robust against choices of multi-judges (e.g., randomly selected judges), indicating that our De-Arena is robust to potential group bias as well~\cite{goel2025great}.

\section{Related Work}

\noindent \textbf{LLM Evaluation and Benchmark.}
Early work~\cite{zhao2019moverscore,zhang2019bertscore,yuan2021bartscore} on evaluating language models primarily focuses on the quality of the generated text, considering the fluency and relevance.
In recent years, large language models~(LLMs) that have undergone pre-training on large-scale corpus, have demonstrated expert-level text generation abilities~\cite{openai2024gpt4ocard, llama3modelcard}, and exhibited strong advanced capabilities~\cite{song2023llmplanner, openai2024o1}, such as reasoning and planning.
Thus, a surge of benchmarks are proposed to assess the multi-aspect capabilities of LLMs, which are either based on closed-ended~\cite {wang2024mmlu,rein2024gpqa} or open-ended questions~\cite{chiang2024chatbot}.
The first type of benchmarks relies on close-ended questions with accurate answers to evaluate LLMs, which simplifies the evaluation process.
However, due to the simple formats of the closed-ended questions, they cannot fully estimate the true user preferences of LLMs in applications~\cite{wu2024rethinking}, and may also be hacked through training on similar data~\cite{sainz-etal-2023-nlp}.
To address this, Chatbot Arena~\cite{chiang2024chatbot} collects open-ended questions, and invites humans to vote on each pair of LLMs based on their outputs.
However, human annotation is costly, and also makes the overall evaluation results difficult to reproduce.
Therefore, a surge of automatic evaluation benchmarks has emerged, employing a strong LLM (e.g., GPT-4~\cite{zheng2023judging,li2023alpacaeval}) or a fine-tuned specialized LLM~\cite{li2023generative,kim2024prometheus} to replace human judges.
Despite the low cost, single-LLM judge-based methods may suffer from the evaluation bias, \eg self-enhancement bias~\cite{zheng2023judging} and verbosity bias~\cite{saito2023verbosity}.
Recent attempts, such as PRD~\cite{li2023prd} and Auto-Arena~\cite{zhao2024auto} have explored multi-LLM judgment to mitigate these issues.
However, using more LLMs as judges would significantly increase the computational and resource cost, limiting the scalability for evaluating a large number of LLMs and new dimensions.

\noindent \textbf{Collective Intelligence.}
Collective intelligence arises when multiple agents collaborate or compete in decentralized networks, often producing more accurate judgments than any single expert alone~\cite{surowiecki2004wisdom,yao2024bayesian}. Research in crowd-based systems~\cite{salminen2015role}, multi-agent systems~\cite{brigui2011multiagent}, and swarm intelligence~\cite{chakraborty2017swarm,gloor2006swarm} shows that collecting diverse perspectives can mitigate individual errors and biases, particularly where no centralized controller is present. For instance, natural examples of ant colonies and bird flocks reveal that complex problems can be addressed effectively through simple agent interactions~\cite{gloor2006swarm}.
In this paper, we extend the principles of collective intelligence to LLM evaluation. We design a large-scale decentralized system in which LLMs simultaneously serve as judges and participants. Because each LLM has a distinct training background, aggregating their judgments reduces the influence of any single model’s biases. Our experiments highlight that evaluation reliability consistently improves with the number of participating models, demonstrating the strong potential of collective intelligence to enable more robust, accurate, and scalable LLM benchmarking.

\begin{figure*}[t]
    \centering
    \includegraphics[width=0.98\linewidth]{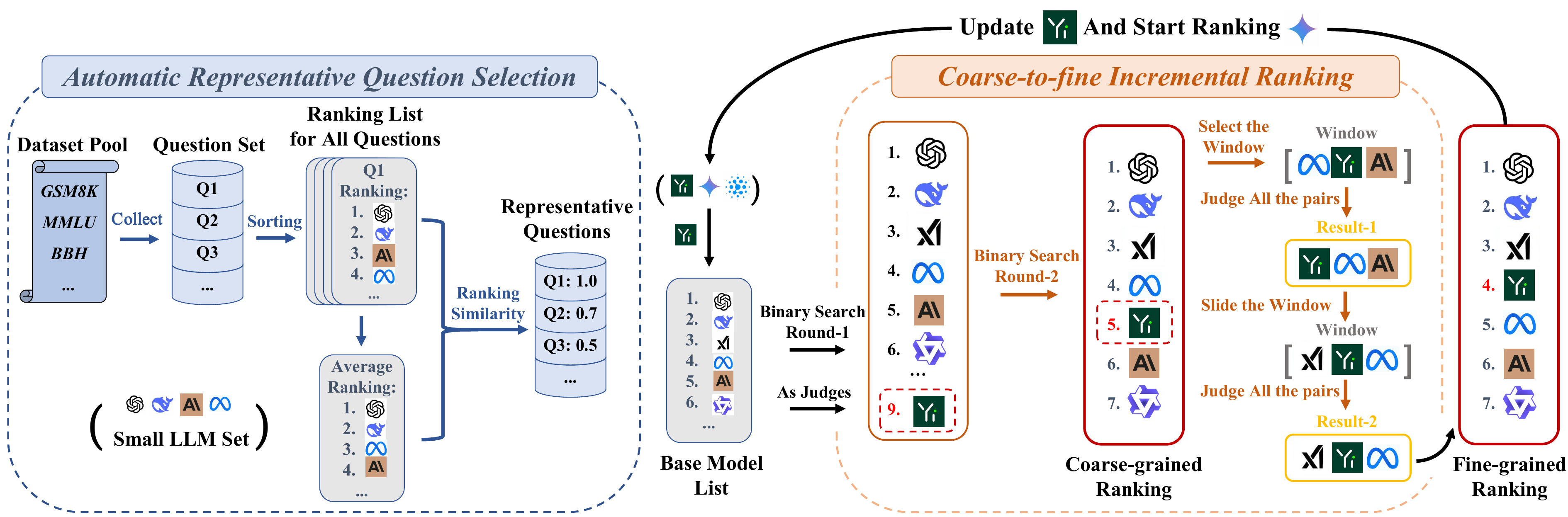}
    \caption{The overview of our method, consisting of the automatic representative question selection strategy (left) and the coarse-to-fine incremental ranking algorithm (right). Here, we show an example that creates a new dimension based on existing open-source datasets, and one of the insert iterations for adding the model Yi into the previous ranking list.}
    \label{fig:main_images}

\end{figure*}

\section{Decentralized Arena}

This section introduces the detailed design of our Decentralized Arena~(De-Arena).
In De-Arena, we focus on the idea of decentralization that uses all LLMs as judges to vote on other model pairs, based on high-quality questions in each dimension. 
It can reduce the cost of gathering human annotations, and also avoid the bias that may arise from relying on a single or a small number of judge models.
To achieve it, we devise the coarse-to-fine incremental sort algorithm to efficiently rank a large number of LLMs, and the automatic question selection algorithm to select representative data for building new evaluation dimension.
The overview of our De-Arena is shown in Figure~\ref{fig:main_images}.

\subsection{Coarse-to-fine Incremental Ranking Algorithm}\label{sec:ranking_algorithm}

Given a set of LLMs $\{m_i\}_{i=1}^{n}$, we aim to sort them into a ranking list $[m_1, \cdots, m_n]$, according to their performance on the collected $k$ questions.
Considering that a surge of stronger LLMs will be developed in the near future, we devise an LLM sort algorithm that supports the incremental insertion of new LLMs into the ranking list.
Concretely, we begin with a small set of ``seed'' models (\ie 6 models), which are ranked using a full-sample pairwise comparison method. In this process, each of the 6 models evaluates and ranks all the other models, excluding itself.
Other models are then incrementally inserted into the rank list, one by one, where all models in the list act as judges to help the new model find its position.
To efficiently insert a new model into the list, we devise the coarse-grained binary search and fine-grained in-window reranking strategies.

\noindent \textbf{Coarse-grained Ranking with Binary Search.}
Given the current ranking list $[m_1, \cdots, m_t]$ with $t$ models, we aim to find the rough position of the $t+1$-th model in an efficient way.
As the ranking list is ordered, we utilize the binary search algorithm~\cite{lin2019binary}, which can quickly narrow down the search space via the logarithmic time complexity.
Concretely, we first compare the new model with the one in the $t/2$-th position of the ranking list, where all other models in the list serve as judges.
Given the collected $k$ questions, all the judge LLMs vote on the outputs of the two models (\ie deciding whose output ``wins'').
If the new LLM owns more wins, we repeat the above step to find the position of the new LLM in the first half list $[m_1, \cdots, m_{t/2-1}]$. Otherwise, we repeat it in $[m_{t/2+1}, \cdots, m_{t}]$.
This loop will continues until narrowing the search space into a certain position, which is the rough position of the model.
The time complexity of this binary search is $\mathcal{O}(kn\log n)$.

\noindent \textbf{Fine-grained In-window Reranking.}
After obtaining the rough position of the new model, we continue to check whether it is suitable and make refinement if necessary.
Here, the new model is compared against its adjacent peers within a defined local window (\eg two models before and after it in the ranking list). The rationale is that these nearby LLMs often own similar capabilities to the new one, whose positions in the ranking list are the hardest to distinguish, warranting closer comparison. 
Concretely, we first compare the new LLM with other in-window models using the collected $k$ questions and rerank them, where all other models outside this window serve as judges. 
If the in-window reranking step leads to a change in the new LLM's position, the process will be repeated within the updated window until the ranking list stabilizes. This functions like a sliding window, guiding the LLM crowd to focus on the most ambiguous comparison pairs, thereby ensuring accurate rankings while significantly reducing computational costs.

\noindent \textbf{Score Generation and Style Control.}
After obtaining the ranking list and pairwise comparison results of all LLM candidates, we follow the methodology used in Chatbot Arena that computes their corresponding Elo score to finalize the ranking results:
\begin{equation}
\small
    R'_{\mathrm{A}}=R_{\mathrm{A}}+K \cdot\left(S_{\mathrm{A}}-\frac{1}{1+10^{\left(R_{\mathrm{B}}-R_{\mathrm{A}}\right) / 400}}\right)
\end{equation}
where $R_A$ is the Elo score of model A that is iteratively updated based on the comparison results, $R'_A$ denotes the updated Elo rating of model A after a pairwise comparison. $S_A$ is a bool value that denotes if model A wins the comparison, and $K$ is the coefficient for the score update. 
As we cannot compare all the model pairs, we follow Chatbot Arena, which uses logistic regression to fit the collected comparison data and estimate the Elo score~\cite{elo1967proposed}.
Here, we consider that the reliability of different LLMs as judges varies. Therefore, we introduce weights in the loss function.
Our rationale is that an LLM with a higher Elo score is more likely to be a qualified judge; hence, we utilize the normalized Elo score as the weight in the loss function. Furthermore, whenever the Elo scores are updated, we dynamically adjust each model's weight based on its new score.
We also follow Chatbot Arena, which incorporates a style control mechanism to reduce the influence of output style.

\subsection{Automatic Representative Questions Selection}\label{sec:question_selection}

To enable scalability in adding arbitrary new evaluation dimensions in De-Arena, we devise an automatic representative question selection algorithm.
To build a new dimension, users only need to collect relevant open-ended questions from open-source datasets.
Then, we utilize the ranking results of LLMs to identify the most representative questions as high-quality examples for evaluating LLMs.

\noindent \textbf{Open-ended Questions Collecting.}
Thanks to the rich open-source datasets in the community, it is easy to search for and collect various relevant open-source datasets for a certain dimension.
However, the collected examples may differ in the formats, \eg, multi-choice and open-ended questions.
In De-Arena, as we can leverage LLMs to compare the outputs of model pairs, we standardize all the data into the open-ended question format by using GPT-4 with an appropriate prompt.

\noindent \textbf{Ranking-based Representative Questions Selection.}
Considering the diverse quality and large scale of the collected questions, we aim to select a few of the most representative ones for testing LLMs.
Instead of randomly sampling, we design a ranking-based method to select questions that lead to consistent ranking lists, ensuring high data quality.
Concretely, for each question $q$ in the collection, we first utilize our ranking algorithm in Section~\ref{sec:ranking_algorithm} to produce the ranking list of a small set of LLMs, denoted as $L$.
Then, we compute the average ranking list for all questions by simply accumulating the position of all LLMs and then sorting them, denoted as $\hat{L}$.
Next, we compute the Spearman correlation $\rho$ between the ranking list of each question and the average list, and use the correlation scores to select the representative questions:
{\small \begin{equation}
    \rho(L, \hat{L})=1-\frac{6 \sum_{i=1}^n (r(L, m_i)-r(\hat{L}, m_i))^2}{n\left(n^2-1\right)}, 
\end{equation}}where $r(L, m)$ returns the position of model $m$ in the list, $n$ is the model number.
Then, questions with higher correlation scores are selected, as they are more capable of representing the ``majority'' preference by yielding ranking results that are highly consistent with the average model rankings.

\section{Experiments}
\label{sec:exp}
\noindent \textbf{Baseline Methods.}
To verify the effectiveness of our approach, we compare it with three types of automatic evaluation benchmarks. Descriptions of each baseline are provided in Appendix~\ref{sec:baseline}, while detailed evaluation settings and implementation details are presented in Appendix~\ref{sec:evaluation_settings} and Appendix~\ref{sec:imp_details}.

$\bullet$ \emph{Closed-ended Datasets based Benchmarks.}
They rely on closed-ended questions to evaluate LLMs.
Since ground-truth answers are provided, they compute answer accuracy to rank LLMs. We select the following benchmarks: (1) \textbf{CompassAcademic}~\cite{2023opencompass}, (2) \textbf{BFCL}~\cite{berkeley-function-calling-leaderboard},
(3) \textbf{Helm Lite}~\cite{liang2022holistic}, (4) \textbf{LiveBench}~\cite{white2024livebench}, (5) \textbf{EQ Bench}~\cite{paech2023eq}, (6) \textbf{MMLU PRO}~\cite{wang2024mmlu}, (7) \textbf{MixEval}~\cite{ni2024mixeval}, (8) \textbf{OpenLLM}~\cite{open-llm-leaderboard-v2}.

$\bullet$ \emph{Single-LLM Judge based Benchmarks.}
In these benchmarks, a single LLM serves as the judge to evaluate LLM pairs. We select the following baselines: (1)\textbf{BiGGen Bench}~\cite{kim2024biggen}, (2)~\textbf{BiGGen Bench (Prometheus 2)}, (3) \textbf{Alpaca Eval 2.0}~\cite{dubois2024length}, (4) \textbf{WildBench}~\cite{lin2024wildbench}.

$\bullet$ \emph{Multi-LLM Judge based Benchmarks.}
In these benchmarks, a set of LLMs is used as the judge models to evaluate LLM pairs. We select the following ones: (1) \textbf{PRD}~\cite{li2023prd}, (2) \textbf{Auto Arena}~\cite{zhao2024auto}.

\begin{table*}
	\caption{Comparison results of the automatic evaluation benchmarks with Chatbot Arena (Spearman Correlation in overall and math dimensions). 
    We report the results from the settings of testing 15, 30, and 66 LLMs. Bold indicates the best results in each group.}
	\small

    \centering
    \begin{tabular}{lccccccc}
    \toprule
    \multirow{2.5}{*}{\textbf{Benchmarks}} & \multicolumn{2}{c}{\textbf{15 LLMs}} & \multicolumn{2}{c}{\textbf{30 LLMs}} & \multicolumn{2}{c}{\textbf{66 LLMs}} & \textbf{Avg.} \\ 
    \cmidrule(lr){2-3} \cmidrule(lr){4-5} \cmidrule(lr){6-7}
    & \textbf{Overall}  & \textbf{Math} & \textbf{Overall}  & \textbf{Math} & \textbf{Overall}  & \textbf{Math} & \\ 
    \midrule
    \textbf{CompassAcademic} & 0.660  & - & 0.674 & - & - & - & - \\
    \textbf{BFCL} & 0.798 & - & 0.813 & - & - & - & - \\
    \textbf{OpenLLM} & 0.892 & - & 0.800 & - & 0.797 & - & - \\
    \textbf{Helm Lite} & 0.725 & 0.656 & 0.748 & 0.660 & 0.750 & 0.665 & 0.701 \\
    \textbf{LiveBench} & 0.906 & 0.900 & 0.920  & 0.913 & 0.925 & 0.916 & 0.913 \\
    \textbf{EQ Bench} & 0.911 & - & 0.860 & - & 0.865 & - & - \\
    \textbf{MMLU PRO} & 0.952 & - & 0.897 & - & 0.897 & - & - \\
    \textbf{MixEval} & 0.954 & - & 0.963 & - & 0.965 & - & - \\
    \midrule
    \textbf{BigGen Bench (Prometheus 2)} & 0.908 & - & 0.924 & - & 0.924 & - & - \\
    \textbf{BigGen Bench} & 0.919 & - & 0.930 & - & 0.931 & - & - \\
    \textbf{Alpaca Eval 2.0} & 0.921 & - & 0.935 & - & 0.935 & - & - \\
    \textbf{WildBench} & 0.894 & 0.917 & 0.907 & 0.932 & 0.910 & 0.934 & 0.916 \\
    \midrule
    \textbf{PRD} & 0.851 & 0.904 & 0.892 & 0.916 & - & - & - \\
    \textbf{Auto Arena} & 0.938 & - & - & - & - & - & - \\
    \midrule
    \textbf{De-Arena} & \textbf{0.957} & \textbf{0.939} & \textbf{0.967} & \textbf{0.952} & \textbf{0.974} & \textbf{0.959} & \textbf{0.958} \\
    \bottomrule   
    \end{tabular}

	\label{tab:eval-table}
\end{table*}



\begin{figure}
    \centering
    \begin{subfigure}[b]{0.44\textwidth}
        \centering
        \includegraphics[width=\linewidth]{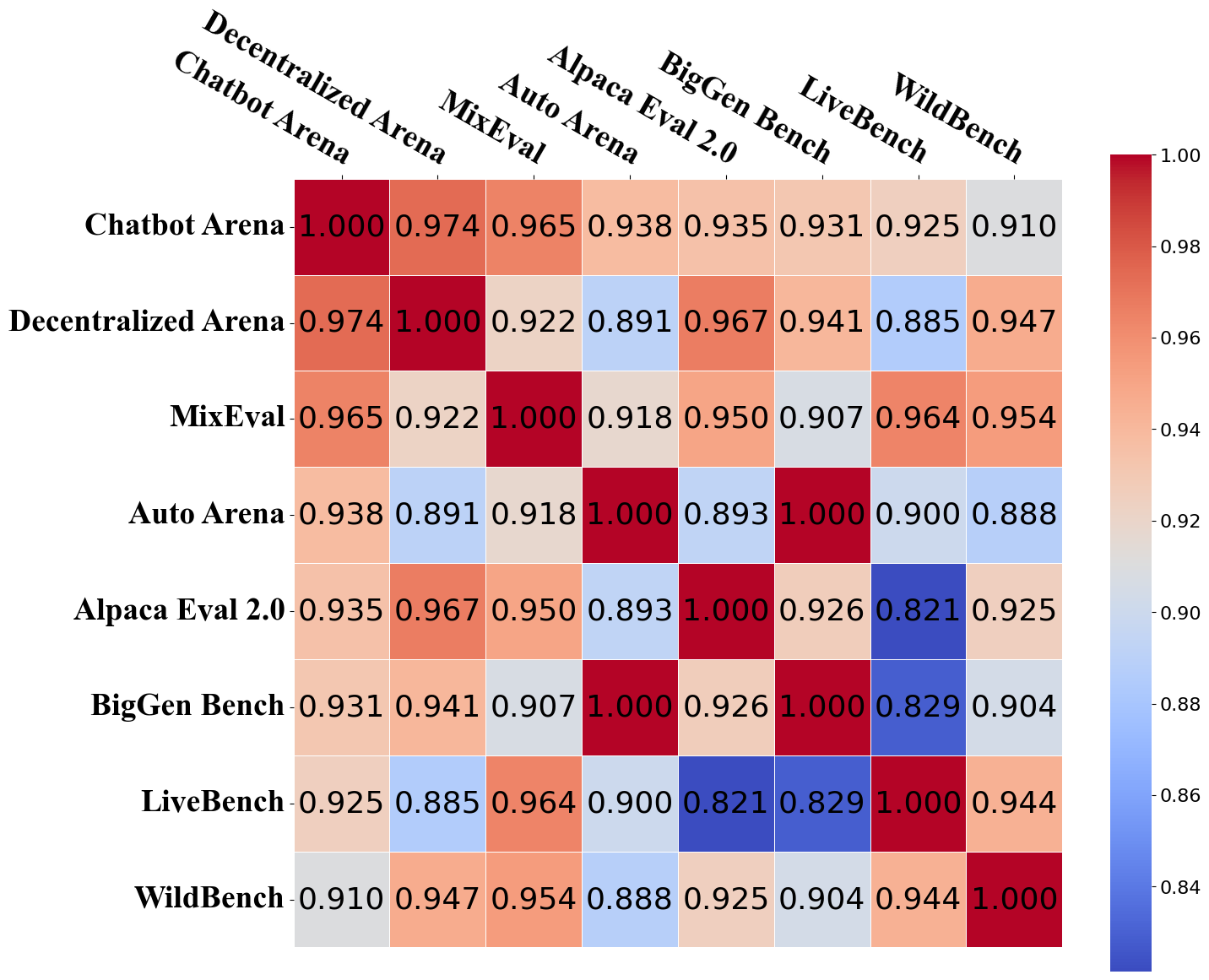}
        \caption{Benchmark Spearman correlation.}
        \label{fig:correl_overall}
    \end{subfigure}
    \hfill
    \begin{subfigure}[b]{0.52\textwidth}
        \centering
        \includegraphics[width=\linewidth]{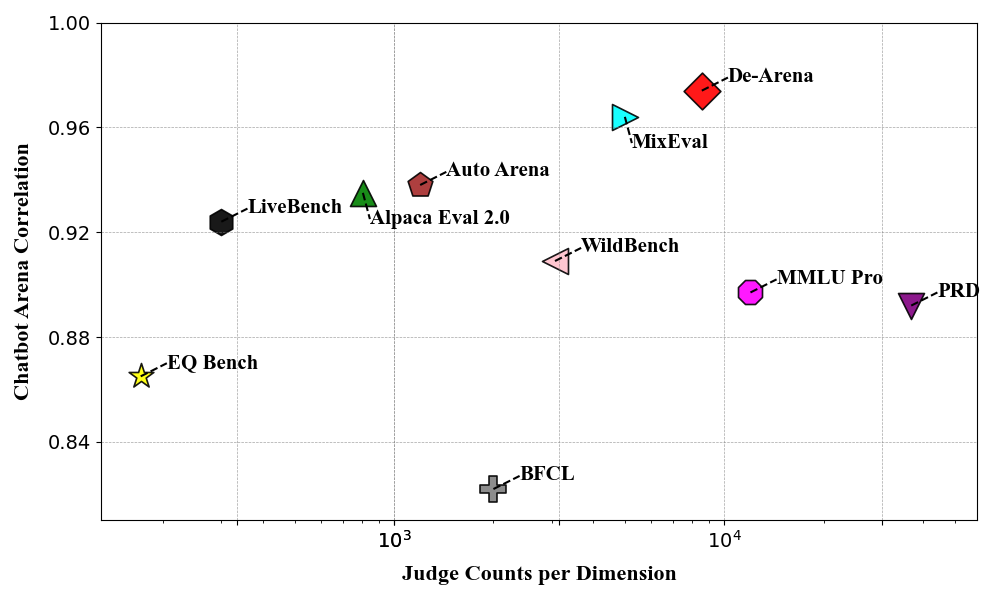}
        \caption{Benchmark cost and performance.}
        \label{fig:cost-study}
    \end{subfigure}

    \caption{(a) Spearman correlation between different LLM benchmarks in the overall dimension. (b)Benchmark cost and performance comparison in the overall dimension, where we show the average judge counts of each model and the correlation with Chatbot Arena.}
    \label{fig:combined-fig}

\end{figure}


\subsection{Results Analysis}

\noindent \textbf{Surpassing Existing Benchmarks.}
The comparison results of different benchmarks are shown in Table~\ref{tab:eval-table}. 
First, we observe that multi-LLM judge-based benchmarks generally perform better than single-LLM judge-based ones. 
This indicates that incorporating multiple LLMs as judges improves the consistency between automated evaluation results and human preferences. Such a way can reduce the preference bias from only one judge model.
Second, by collecting high-quality questions for evaluation, MixEval and WildBench outperform other closed-ended and single-LLM judge-based benchmarks, respectively.
MixEval carefully controls its query distribution to match with real-world user queries, while WildBench collects massive real-world tasks (\ie 1024).
It demonstrates the importance of selecting proper datasets for evaluation.
Besides, our De-Arena surpasses all baselines in most evaluation settings and the average value.
In De-Arena, we extend the multi-LLM judge strategy into a more democratic paradigm where all LLMs are both the judges and are to be evaluated, to further reduce the bias.
Furthermore, we devise an automatic question selection strategy that leverages the correlation of ranking results to find the most representative questions for evaluation.
The above strategies greatly improve the reliability and scalability of our method, in contrast to baselines that require human involvement and a large amount of data.

In addition, with the increasing of test LLM number, the difficulty of accurately ranking all LLMs also increases.
As a result, the correlation scores of most baselines with Chatbot Arena have also decreased.
Here, we can see that our De-Arena can achieve a stable performance and even perform better in the math dimension.
The reason is that the involvement of more LLMs also introduces more judge models, which can reduce the bias caused by a few judges, further improving the reliability. We also report the correlation between our De-Arena and other best-performing six benchmarks in Figure~\ref{fig:correl_overall}.
We can see that our De-Arena always has a high correlation with all the benchmarks, \ie $>0.85$.
It further indicates the effectiveness of our method for producing reliable ranking results, as existing benchmarks, echoing with the superior performance in Table~\ref{tab:eval-table}.

\begin{wraptable}[12]{r}{0.55\textwidth}
    \centering
    \small

    \caption{Judge methods vs. Chatbot Arena: Correlation ($\uparrow$) and Rank Difference ($\downarrow$).}
    \begin{tabular}{@{}lcccc@{}}
        \toprule
        \multirow{2}{*}{\textbf{Methods}} & \multicolumn{2}{c}{\textbf{MT-Bench}} & \multicolumn{2}{c}{\textbf{Math}} \\ 
        \cmidrule(lr){2-3} \cmidrule(lr){4-5}
        & \textbf{Corr$\uparrow$} & \textbf{R-Diff$\downarrow$} & \textbf{Corr$\uparrow$} & \textbf{R-Diff$\downarrow$} \\ 
        \midrule
        LLaMA-3-70B & 0.815 & 1.71 & 0.934 & 1.14 \\
        Gemma-2-27B & 0.930 & 1.29 & 0.932 & 1.00 \\
        Qwen2-70B & 0.938 & 1.00 & 0.942 & 1.14 \\
        \midrule
        De-Arena & \textbf{0.956} & 1.00 & \textbf{0.952} & 1.00 \\
        \bottomrule   
    \end{tabular}
    \label{tab:bias-study}
\end{wraptable}

\noindent \textbf{Reducing Single-Judge Biases.}
In De-Arena, our major contribution is to utilize all LLMs as judges to democratically vote all the model pairs, reducing the single-judge evaluation bias and improving reliability.
To study it, we compare our De-Arena with several of its variations using a single LLM as the judge, including LLaMA-3-70B, Gemma-2-27B, Qwen2-70B-inst, GPT-4o-2024-08-06. Here, we report the Spearman correlation and the difference in the ranked LLMs between all methods with Chatbot Arena.
As presented in Table~\ref{tab:bias-study}, the performance of our De-Arena is consistently better than all other variations, with higher Spearman correlation and lower rank difference.
It indicates that our democratic voting strategy can avoid the ranking results being biased by the preference of few LLMs.
Also, in our case study, we find that LLaMA-3-70B and Gemma-2-27B are prone to vote for themselves and the same series of LLMs, causing their ranks to rise drastically.

\begin{wrapfigure}[15]{R}{0.39\textwidth}

	\begin{center}
        \includegraphics[width=1\linewidth]{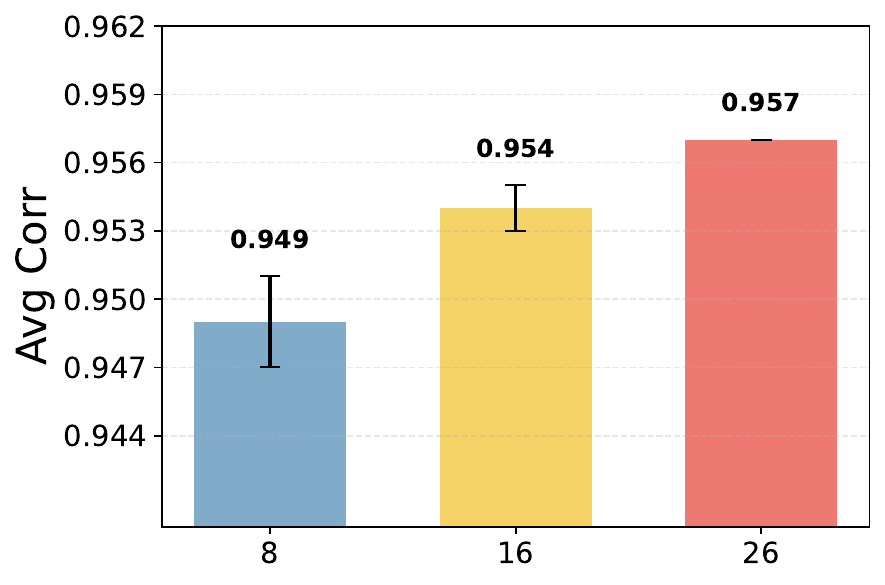}
	\end{center} 

    \caption{Spearman correlation with Chatbot Arena across varying judge model number.}
	\label{fig:judge_model_num} 

\end{wrapfigure}



\noindent \textbf{Robustness against Potential Group Biases.} 
Since De-Arena leverages the collective intelligence of LLMs to judge each other, it effectively mitigates single-judge biases. To further assess its robustness and potential group biases, we varied the number of judge models across three settings: 8, 16, and 26. For the 8 and 16 settings, we randomly sampled five different judge sets to evaluate stability; for 26, we used all suitable open-source models. The final outcomes are presented in Figure~\ref{fig:judge_model_num}. We observe that the performance consistently improves as the number of judge models increases, with the best performance achieved when the number reaches 26. This can be attributed to the fact that a larger number of judge models enables a more democratic and decentralized evaluation process. As the judge pool grows, the collective intelligence effect becomes more pronounced, which helps to further mitigate the biases of individual models. Meanwhile, since the group consists of highly diverse models, group biases are minimally introduced. This highlights the strong robustness of using multiple judges in the evaluation process.


\noindent \textbf{Cost and Scalability Study.}
In De-Arena, we devise the coarse-to-fine ranking algorithm and question selection strategy to reduce the cost of scaling the LLM number.
To evaluate this efficiency, we estimate the cost of our De-Arena with other benchmarks for comparison.
As it is hard to compute the detailed cost, we count the average comparison number of each LLM, which is relevant to the number of test questions and voting counts.
As shown in Figure~\ref{fig:cost-study}, our De-Arena achieves the best performance among all benchmarks, with slightly higher cost than single-LLM judge-based ones.
The reason is that we employ the representative question selection strategy to reduce the number of test questions (\eg 100 instances), and also the ranking algorithm to reduce the voting counts of judge models.
The above designs greatly reduce the cost from both sides.
Besides, as De-Arena has the lower cost and higher correlation with Chatbot Arena, it shows strong potential as an effective and scalable automatic counterpart for broader real-world applications.

\begin{figure}[htbp]
    \centering
    \begin{subfigure}[b]{0.62\textwidth}
        \centering
        \includegraphics[width=\linewidth]{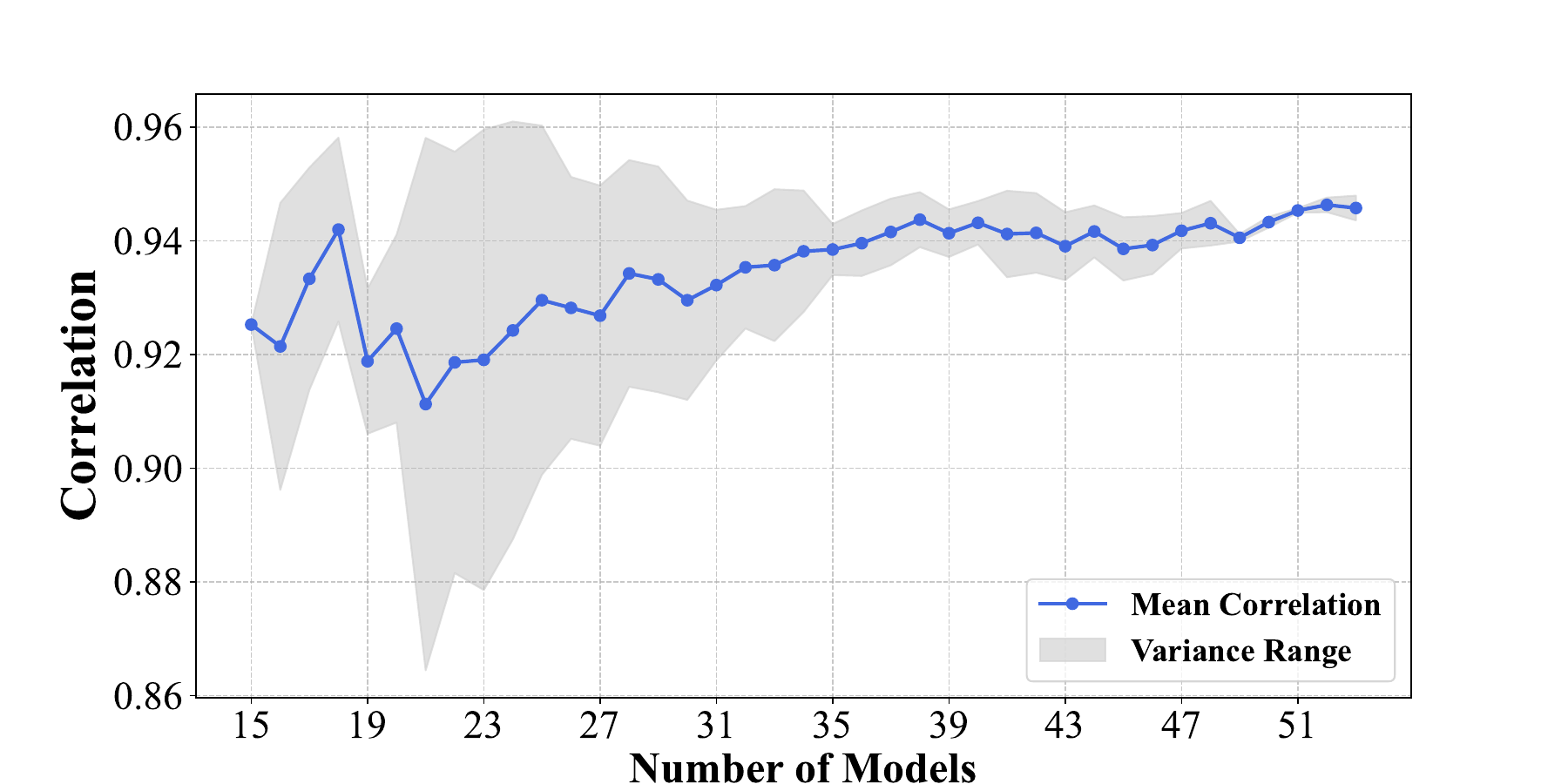}
        \caption{De-Arena Correlation Curve}
        \label{fig:convergence}
    \end{subfigure}
    \hfill
    \begin{subfigure}[b]{0.37\textwidth}
        \centering
        \includegraphics[width=\linewidth]{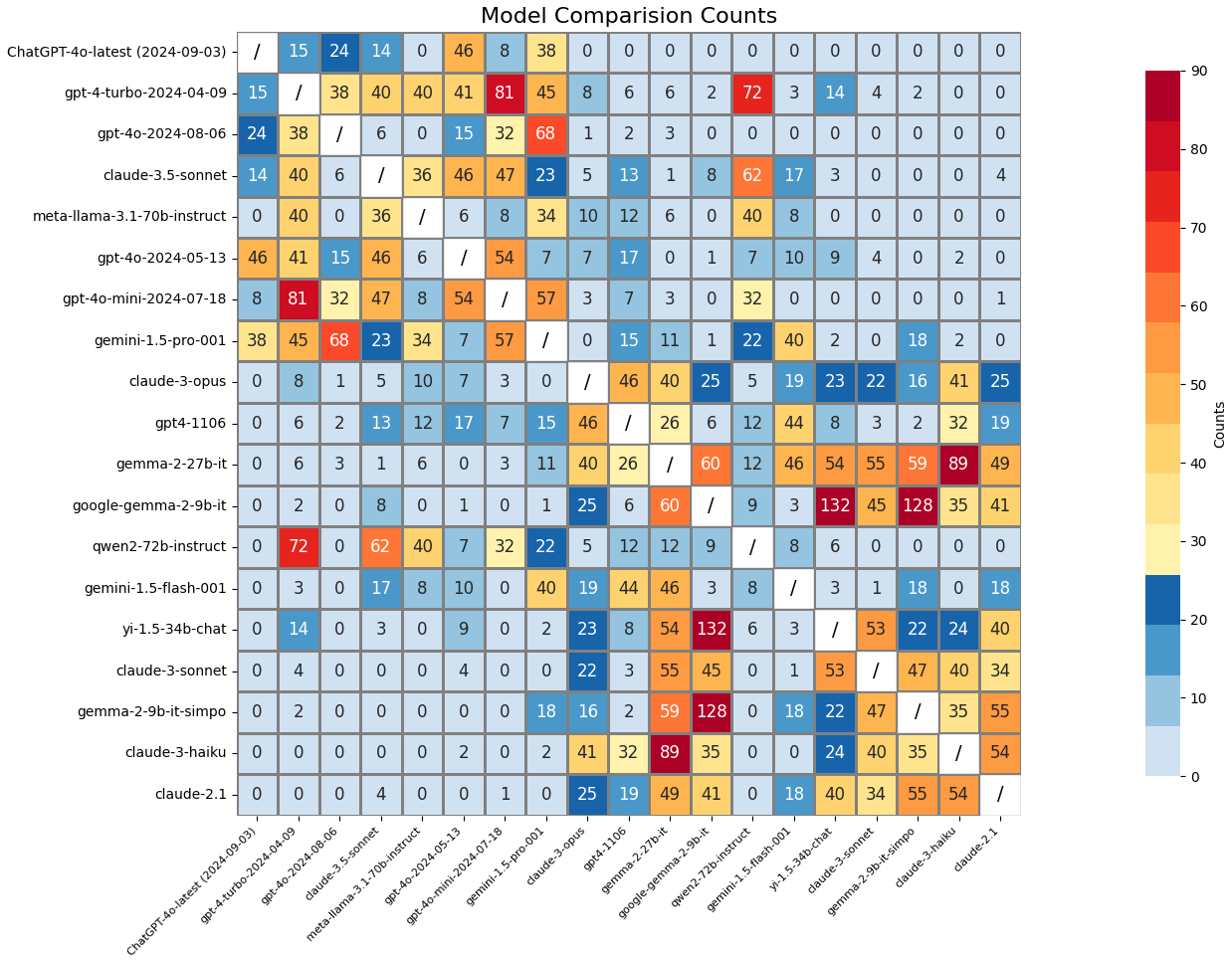}
        \caption{LLM comparison counts.}
        \label{fig:compare_count_map}
    \end{subfigure}
    \caption{(a) De-Arena's mean (blue curve) and variance (shaded area) of Chatbot Arena correlation in the MT-bench dimension, with the increase in LLM number. (b) The distribution map of the LLM comparison counts in the MT-Bench dimension.}
\end{figure}

\noindent \textbf{Convergence Study.}
As our De-Arena adopts the coarse-to-fine incremental ranking strategy, the insertion order of LLMs might affect the stability of the final ranking results.
To study it, we run our method five times, using different random seeds to shuffle the insertion order, and compute the mean and variance correlations with Chatbot Arena, in the MT-bench dimension.
As shown in Figure~\ref{fig:convergence}, we can see that with the involvement of more models, our ranking results become more stable and robust with higher correlation and lower variance.
It demonstrates the scalability of our decentralized evaluation strategy with the scaling of LLMs.
The more models participate in the evaluation process, the more reliable and trustworthy the ranking results for all models.

\subsection{Ablation and Variation Study}
\begin{table}[t]
    \centering
    \small
    \caption{Correlations~(Corr) and rank differences~(R-Diff) between different judge methods and Chatbot Arena in MT-Bench and Math dimensions. $\uparrow$ denotes the higher the better, while $\downarrow$ denotes the lower the better.}
    \begin{tabular}{lcccc}
        \toprule
        \multirow{2}{*}{\textbf{Methods}} & \multicolumn{2}{c}{\textbf{MT-Bench}} & \multicolumn{2}{c}{\textbf{Math}} \\ 
        \cmidrule(lr){2-3} \cmidrule(lr){4-5}
        & \textbf{Corr$\uparrow$}  & \textbf{Judges$\downarrow$} & \textbf{Corr$\uparrow$}  & \textbf{Judges$\downarrow$} \\ 
        \midrule
        Ours & \textbf{0.957} & 521495 & \textbf{0.962} & 808074 \\ 
        - w/o Fine & 0.952 & 320715 & 0.961 & 489741 \\ 
        - w/o Coarse & 0.954 & 352245 & 0.959 & 520580  \\ 
        - Full Sample & 0.956 & 2245874 & \textbf{0.962} & 3660355 \\ 
        \bottomrule   
    \end{tabular}
    \label{tab:ablation-Ranking}
\end{table}

\noindent \textbf{Coarse-to-fine Ranking Algorithm.}
To study the effectiveness of our coarse-to-fine ranking algorithm, we remove the coarse-grained binary search ranking and fine-grained in-window reranking strategies, to build two variations for comparison: (1) \textbf{w/o Coarse}: ours without coarse-grained binary search; (2) \textbf{w/o Fine}: ours without fine-grained in-window reranking. 
Besides, we also built the variation that uses all LLMs to vote all the LLM pairs, namely (3) \textbf{Full Ranking}: ours with full ranking.
We conduct the experiments on MT-Bench and Math dimensions, and report the correlation with Chatbot Arena and the number of judges.
As shown in Table~\ref{tab:ablation-Ranking}, among all the variations, our De-Arena can well balance the performance and the cost.
In De-Arena, both the coarse-grained binary search and the fine-grained in-window reranking algorithms contribute to performance improvement while only slightly increasing the number of judge.
Without these strategies, the variation that needs to fully rank all model pairs greatly increases the cost ($\times 4$ judge counts).



\begin{table}[t]
    \centering
    \begin{minipage}[t]{0.52\textwidth}
        \centering
        \small
	    \caption{Stability study of our coarse-to-fine ranking algorithm. In each dimension, we conduct five random experiments by shuffling the insertion order of models.}
        \resizebox{\textwidth}{!}{%
    	\begin{tabular}{lcccc}
    		\toprule
    		\textbf{Dimension} & MT-Bench &  Algerba & Geometry & Probability\\ 
    		\midrule
    		Avg Corr & 0.957 & 0.942 & 0.956 & 0.961 \\
            Std & 0.0019 & 0.0021 & 0.0015 & 0.0016\\
    		\bottomrule   
    	\end{tabular}
    	\label{tab:error-bar}
        }
    \end{minipage}
    \hfill
    \begin{minipage}[t]{0.47\textwidth}
        \centering
        \small
    	\caption{Stability study under variations of our coarse-to-fine ranking algorithm. Each dimension, we conduct five random experiments by shuffling the insertion order of models.}
        \resizebox{\textwidth}{!}{%
    	\begin{tabular}{lcccc}
    		\toprule
    		\textbf{Methods} & Ours &  - w/o Fine & - w/o Coarse\\
            \midrule
    		Avg Corr & 0.957 & 0.953 & 0.955 \\
            Std & 0.0019 & 0.0031 & 0.0029 \\
    		\bottomrule   
    	\end{tabular}
    	\label{tab:error-bar-var}}
    \end{minipage}
\end{table}

\noindent \textbf{Stability Study of Coarse-to-fine Ranking Algorithm}\label{sec:Stability_Study}
In De-Arena, we adopt an incremental insertion approach, where models are inserted in different orders and ranked accordingly. To better demonstrate the stability of our algorithm, we conduct experiments showing that the insertion order of models into De-Arena has minimal impact on the final ranking results.
Specifically, we perform five random shuffles of the model insertion order and compute the Spearman correlation between the resulting rankings and those from Chatbot Arena. As shown in Table~\ref{tab:error-bar}, the correlation remains consistently high with very low standard deviation, confirming the robustness and stability of the coarse-to-fine ranking algorithm. Furthermore, to verify the stability of our algorithm's variations, we conduct the same randomized insertion experiments under MT-Bench. As shown in Table~\ref{tab:error-bar-var}, although the variations do not perform as well as the full algorithm, their standard deviations remain very low, indicating strong stability.
Therefore, the full coarse-to-fine ranking algorithm demonstrates the best performance and is well-suited for scenarios that require highly accurate and stable rankings.


\begin{figure}[htbp]
    \centering
    \begin{subfigure}[b]{0.44\textwidth}
        \centering
        \includegraphics[width=\linewidth]{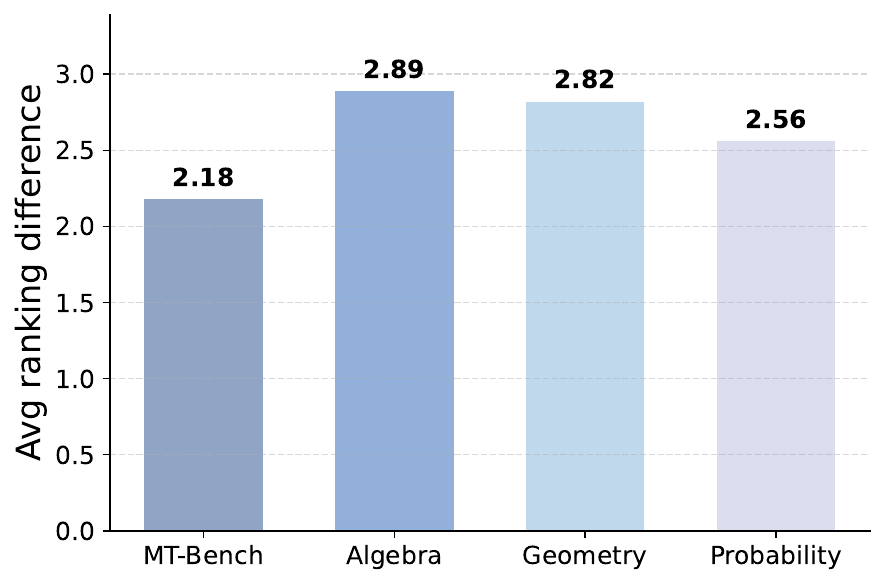}
        \caption{Ranking difference.}
      	\label{fig:binary_ranking_dif} 
    \end{subfigure}
    \hfill
    \begin{subfigure}[b]{0.51\textwidth}
        \centering
        \includegraphics[width=\linewidth]{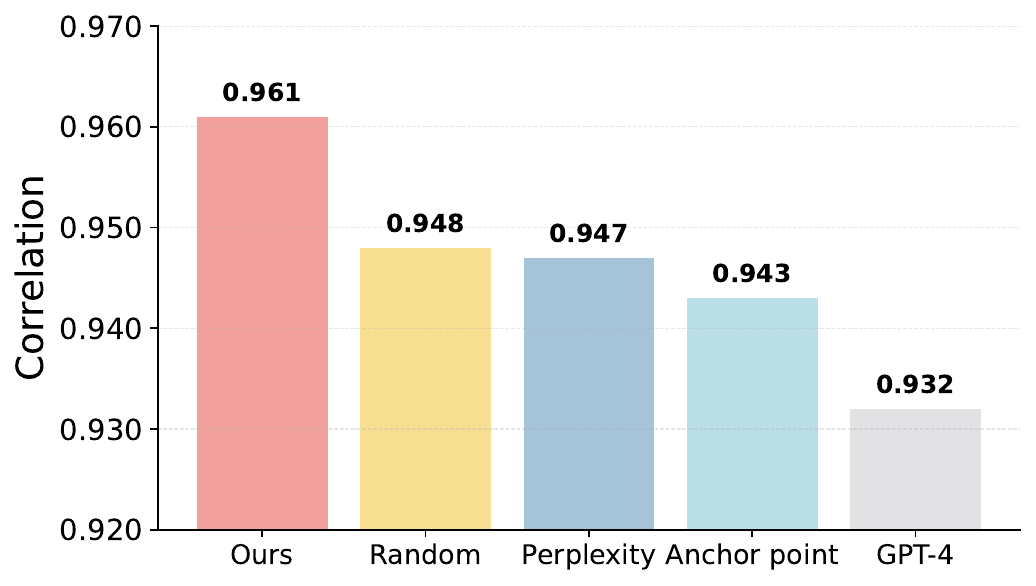}
        \caption{Correlation across question selection methods.}
    	\label{fig:queries_selection_table} 
    \end{subfigure}
    \caption{(a) Binary search ranking differences between binary search and ground truth in four dimensions. (b) Spearman correlation with Chatbot Arena for different question selection methods.}
\end{figure}


\noindent \textbf{Accuracy Study of Coarse-grained Binary Search}
De-Arena heavily relies on the Coarse-to-Fine Ranking Algorithm, with the accuracy of the binary search in the first step being crucial for identifying the approximate ranking range of models. To better demonstrate the accuracy of the binary search, we monitor the absolute difference between the binary search ranking and the ground truth ranking during the insertion process across four dimensions. Finally, we compute the average ranking difference for all models. As the results shown in Figure~\ref{fig:binary_ranking_dif}, the coarse-grained binary search ranking for each model across all dimensions is very close to its ground truth ranking. With the subsequent fine-grained ranking adjustments, the accuracy of the rankings is further improved.




\noindent \textbf{Question Selection Algorithm.}
To evaluate the effectiveness of our representative question selection algorithm, we compare it with several variations using different strategies: (1) \textbf{Random} that randomly selects the questions; (2) \textbf{Perplexity} that uses the perplexity of LLaMA-3-8B~\cite{llama3modelcard} to rank and select the top ones; (3) \textbf{Anchor point}~\cite{vivek-etal-2024-anchor} that selects an optimal subset of sub-problems to represent the full dataset; (4) \textbf{GPT-4} that crafts prompts to guide GPT-4 to rank and select the top ones.
Here, we utilize them to select 32 questions from the 80 questions in MT-Bench dataset.
As shown in Figure~\ref{fig:queries_selection_table}, all the variations perform not better than De-Arena, indicating the effectiveness of our question selection algorithm.
Here, random is a robust baseline that can outperform other variations, while our methods can lead to a higher correlation with Chatbot Arena than it. 
The reason is that we focus on selecting the most representative questions that reflect the majority, based on ranking similarity. This approach effectively identifies the most useful ones for testing LLMs.


\begin{wraptable}[8]{r}{0.50\textwidth}
	\caption{Ablation study results about the Elo weight in different dimensions.}
	\small
	\centering
	\begin{tabular}{lcc}
		\toprule
		\textbf{Dimension} & \textbf{with Elo Weights} & \textbf{No Weights} \\ 
		\midrule
		MT-Bench & \textbf{0.957} & 0.949  \\
            Math & \textbf{0.959} & 0.953 \\
		\bottomrule   
	\end{tabular}
	\label{tab:weight-ablation}
\end{wraptable}

\noindent \textbf{Weights for Judge Models.}
In De-Arena, we consider that different models have varying abilities for judging LLMs, thereby devising the weighting mechanism. It aims to assign higher weights for stronger models and lower weights for weaker ones.
To study its effectiveness, we remove it and compare the performance changes in the MT-Bench and three math sub-dimensions.
As shown in Table~\ref{tab:weight-ablation}, removing the weights would lead to performance degradation in the correlation score. It indicates the effectiveness of the weighting mechanism we implemented.



\noindent \textbf{Comparison-count Distribution.}
Our De-Arena adopts the coarse-to-fine ranking algorithm, which can allocate more comparisons on the hard-to-distinguish LLM pairs with neighboring positions in the ranking list.
To study it, we visualize the comparison-count distribution for all LLMs in Figure~\ref{fig:compare_count_map}.
We can observe that the collective LLM intelligence automatically focuses primarily on the neighboring LLM pairs (those close to the diagonal), which are also equivalent to those with near 50\% win rates in Figure~\ref{fig:winrate_map}.
In contrast, comparisons between LLMs with large performance gaps are sparse (or even omitted), reducing the overall computation cost.
Such a distribution is all thanks to our ranking algorithm, where the binary search and in-window reranking help reduce the unnecessary comparisons with predictable results and concentrate on the ambiguous pairs.

\section{Conclusion}

In this paper, we propose Decentralized Arena~(De-Arena), a democratic and fully automatic LLM evaluation system where the models to be evaluated can also evaluate each other.
To make it a more efficient and automatic system, we devised the coarse-to-fine incremental ranking and representative question selection strategies. These innovations enable De-Arena to scale effectively to a large number of LLMs and support evaluation across fine-grained, diverse dimensions
Extensive experiments have verified the reliability and scalability of our De-Arena.
In the future, we will extend our De-Arena by including more LLMs and useful evaluation dimensions, supporting fully automatic new dimension discovery and evaluation, and further exploring the evaluation of super-human intelligence.

\bibliographystyle{unsrt}
\bibliography{ref.bib}


\appendix

\newpage
\appendix
\begin{table}[t]
	\caption{Style control ablation study results across different dimensions. Each column represents a different style control method applied, and All denotes controlling all of them.}
	\small
	\centering
	\begin{tabular}{lccccc}
		\toprule
		\textbf{Dimension} & \textbf{Length} & \textbf{Header} & \textbf{List} & \textbf{Bold} & \textbf{All} \\ 
		\midrule
		MT-Bench & \textbf{0.933} & 0.914 & 0.911 & 0.897 & 0.932 \\
		Algebra & 0.909 & 0.909 & 0.909 & \textbf{0.911} & 0.906 \\
		Probability & 0.934 & 0.930 & 0.929 & 0.929 & \textbf{0.935} \\
		Geometry & 0.932 & \textbf{0.934} & 0.933 & 0.932 & 0.924 \\
		\bottomrule   
	\end{tabular}
	\label{tab:style-control}
\end{table}

\section{Style Control Study}
Due to variations in training data, different models often exhibit distinct output styles. Following the approach of Chatbot Arena, we defined four styles (\ie Length, Header, List, Bold) and calculated the correlation between these styles and the Chatbot Arena style control. Also, we add the results by controlling all the them.

As the results shown in Table~\ref{tab:style-control}, we can see that controlling all of the styles can achieve better performance in all these dimensions.
In contrast, only using one of them would have an improvement on a certain dimension, but might also affect the performance in other ones.
It indicates that controlling all styles is capable of well balancing the capability in all evaluation aspects.

\begin{table}[t]
	\caption{Hyperparameter tuning results about window size.}
	\small
	\centering
	\begin{tabular}{lcccc}
		\toprule
        \multirow{2.5}{*}{\textbf{Window size}} & \multicolumn{2}{c}{\textbf{MT-Bench}} & \multicolumn{2}{c}{\textbf{Math}} \\ 
		\cmidrule{2-5}
		& \textbf{Corr$\uparrow$}  & \textbf{Judges$\downarrow$} & \textbf{Corr$\uparrow$}  & \textbf{Judges$\downarrow$} \\ 
		\midrule
		\multicolumn{1}{c}{1} & \textbf{0.957} & 521495 & \textbf{0.962} & 808074 \\ 
		\multicolumn{1}{c}{2} & 0.953 & 684460 & 0.960 & 1069615  \\ 
		\multicolumn{1}{c}{3} & 0.955 & 892260 & 0.960 & 1284068 \\ 
		\bottomrule   
	\end{tabular}
	\label{tab:window}
\end{table}

\begin{table}[t]
	\caption{Hyperparameter tuning results about base model number.}
	\small
	\centering
	\begin{tabular}{lcccc}
		\toprule
		\textbf{Base Model Number} &\textbf{3} &  \textbf{6} & \textbf{9} &  \textbf{12}\\ 
		\midrule
		MT-Bench & 0.948 & \textbf{0.957} & 0.952 & 0.954 \\
        Math & 0.961 & \textbf{0.962} & 0.960 & 0.960 \\
		\bottomrule   
	\end{tabular}
	\label{tab:base-model-num}
\end{table}

\section{Hyper-parameter Tuning}
\label{sec:hyper-parameter}
In De-Arena, the window size and base model number are two hyper-parameters that control the cost of in-window reranking and the initial ranking list, respectively.
Here, we study their best settings by varying them in $[1,2,3]$ and $[3,6,9,12]$, respectively.
As the results shown in Table~\ref{tab:window}, setting the window size to 1 can lead to the fewest judge counts, and also achieve a good correlation score, which well balances the performance and the cost.
As shown in Table~\ref{tab:base-model-num}, we can observe that using 6 base models can achieve the best performance.
The reason is that too few or too many models would cause the instability of the ranking list during incrementally inserting new models.

\section{Impact Statement}
\label{sec:impact_statement}
Our De-Arena aspires to reshape LLM benchmarking by harnessing collective intelligence rather than relying on few ``authority'' models or costly human annotation. It may carry several potential ethical and societal impacts:

$\bullet$ By involving every participating LLM as both evaluator and evaluated, De-Arena aims to reduce single-model dominance and mitigate systemic biases. It encourages more equitable participation and transparent performance comparisons, fostering an environment in which models from diverse teams—industry, academia, or open-source communities—can be assessed on a level playing field.

$\bullet$ Traditional benchmarking often depends on extensive human annotation, which can be labor-intensive, subjective, and slow. De-Arena’s automatic evaluation minimizes human oversight and lowers costs, potentially democratizing access to robust evaluation for smaller research groups or underfunded institutions and easing the ethical burden associated with human annotators’ time and well-being.

$\bullet$ In contrast to single-judge approaches, a decentralized, multi-LLM system spreads accountability across many models. When combined with transparency about each model’s contributions to a final ranking, the system can better highlight disagreements or harmful biases among models. This collective responsibility promotes more nuanced scrutiny of anomalies or potentially harmful content.

$\bullet$ While distributing decision-making reduces reliance on any single model’s biases, emergent group biases can still arise if many models share similar training data or user bases. Continued research is needed to detect and mitigate these collective distortions, especially for underrepresented languages or cultural contexts.

\begin{promptbox}[Questions with the Higher and Lower Scores]{lightgreen}
\textbf{Selected Question with Higher Score:}

$\bullet$ You have been tasked with designing a solar-powered water heating system for a residential building. Describe the key components and considerations you would include in your design. Design a five-step workflow.\\

\textbf{Unselected Question with Lower Score:}

$\bullet$ What is the central dogma of molecular biology? What processes are involved? Who named this?

\end{promptbox}

\section{Case Study for Question Selection}
To better show the effectiveness of our representative question selection algorithm, we show the questions with the higher and lower scores using our method in the above example.
We can observe that the selected question with the higher score is indeed with higher quality. It contains more detailed task description and requires multiple special knowledge to solve it.
In contrast, for the question with the lower score, its required knowledge is relatively limited. As no clear instruction is given, it is not easy to distinguish the quality of the potential outputs from LLMs.

\begin{table}[t]
	\caption{Comparison of the ranking results using LLaMA-3-70B as the judge and our method, respectively.}
	\small
	\centering
	\begin{tabular}{c|c}
		\toprule
		\textbf{Results from De-Arena} & \textbf{Results using LLaMA-3-70B} \\ 
		\midrule
		o1-mini & llama-3-70b-instruct \\
        o1-preview & meta-llama-3.3-70b-instruct \\
        gpt-4o-2024-05-13 & o1-mini \\
        meta-llama-3.3-70b-instruct & gpt-4o-2024-08-06 \\
        gpt-4o-2024-08-06 & o1-preview \\
        qwen2-72b-instruct & gpt-4o-2024-05-13\\
        gemma-2-27b-it & qwen2-72b-instruct \\
        gemma-2-2b-it & gemma-2-27b-it \\
        llama-3-70b-instruct & gemma-2-2b-it \\
        gemma-1.1-7b-it & gemma-1.1-7b-it \\
        gemma-1.1-2b-it & gemma-1.1-2b-it\\
        qwen2.5-1.5b & llama2-7b-chat\\
        llama2-7b-chat & llama2-13b-chat \\
        llama2-13b-chat & qwen2.5-1.5b \\
        qwen1.5-4b-chat& qwen1.5-4b-chat \\
		\bottomrule   
	\end{tabular}
	\label{tab:bias-rank}
\end{table}

\section{Case Study for Ranking Results}
To study the ranking bias in single-LLM judge based methods and our approach, we show the ranking results using only LLaMA-3-70B as the judge and our De-Arena in Table~\ref{tab:bias-rank}.
In the ranking list using LLaMA-3-70B as the judge, LLaMA-3-70B itself and its fine-tuned version Meta-LLaMA-3.3-70B-instruct are both ranked into the first and second positions, respectively.
It demonstrates the existence of the evaluation bias for single-LLM judge based methods.
In contrast, our De-Arena can produce a more reliable ranking results, which is more consistent as human preference (as shown in Table~\ref{tab:bias-study}).
It demonstrates that using more LLMs as judges is promising to obtain more reliable ranking results than using one or few judge models.

\section{Fine-grained Dimension Correlation}
\begin{figure*}[!htbp]
    \centering
    \includegraphics[width=0.9\linewidth]{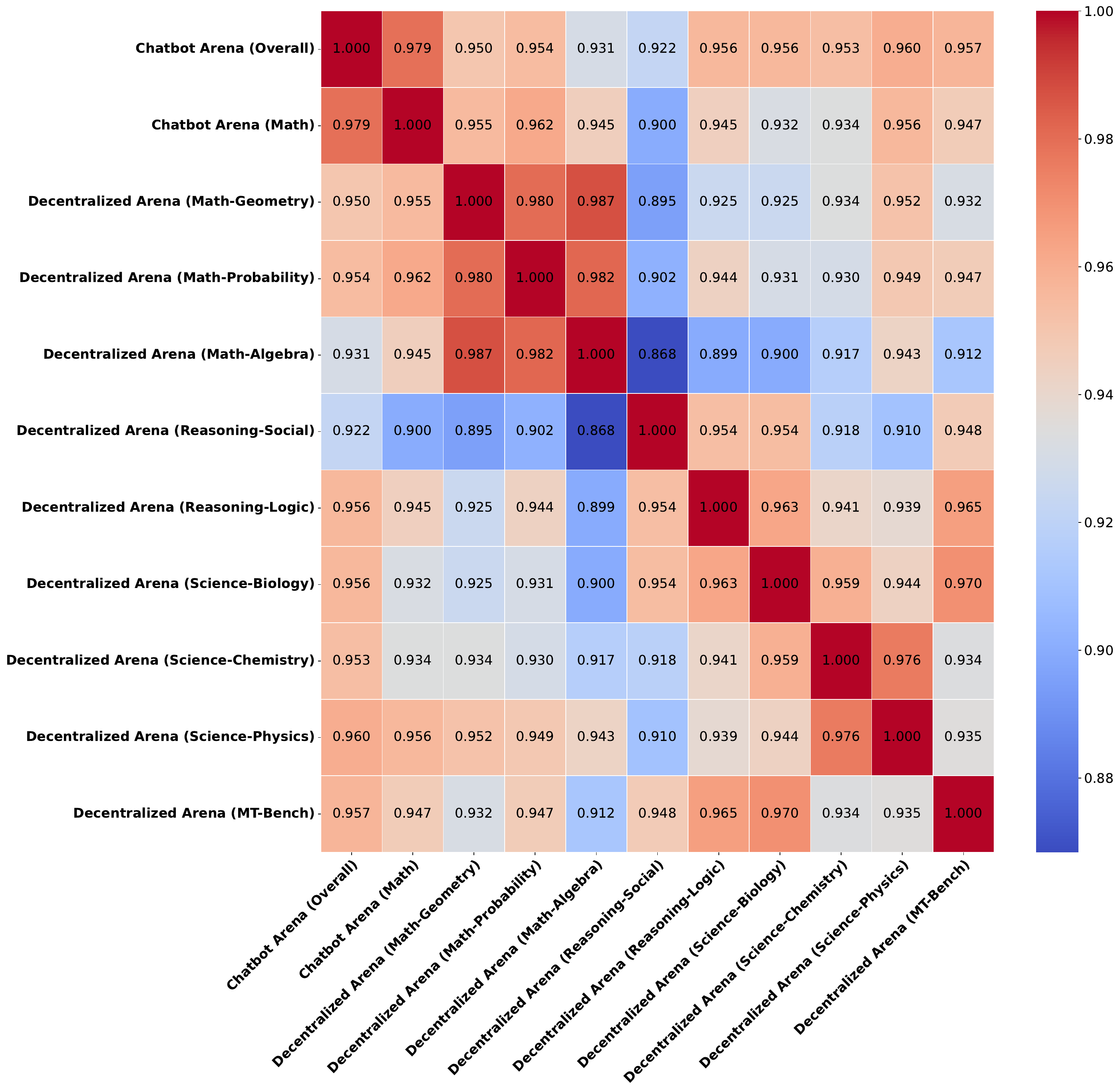}
    \caption{Correlations between the ranking results from different dimensions in De-Arena and Chatbot Arena.}
    \label{fig:correl_dimensions}
\end{figure*}

Our approach achieves high correlations with human judge based Chatbot Arena (95\% in the ``Overall'' dimension).
Here, we further report the correlation between each dimension from our De-Arena and the dimensions from Chatbot Arena (\ie Overall and Math) in Figure~\ref{fig:correl_dimensions}.
We can see that the correlation scores are always high across these dimensions ($>0.85$), indicating the consistency of our automatic ranking results and human preference.
For the fine-grained sub-dimensions about a certain capability (\ie math, reasoning, and science), their correlations are also relatively higher.

\section{Baseline Details}\label{sec:baseline}
$\bullet$ \emph{Closed-ended Datasets based Benchmarks.}

(1) \textbf{CompassAcademic}~\cite{2023opencompass} selects a set of open-source datasets and benchmarks and integrates them to evaluate LLMs.
(2) \textbf{BFCL}~\cite{berkeley-function-calling-leaderboard} evaluates LLMs' ability to accurately call functions in real-world data.
(3) \textbf{Helm Lite}~\cite{liang2022holistic} is a lightweight benchmark consisting of nine scenarios, including math reasoning, medical QA, and long context QA.
(4) \textbf{LiveBench}~\cite{white2024livebench} contains 18 diverse tasks across 6 categories, which minimizes potential contamination by releasing new questions monthly.
(5) \textbf{EQ Bench}~\cite{paech2023eq} is an emotional intelligence benchmark for evaluating LLMs' ability to understand complex emotions and social interactions.
(6) \textbf{MMLU PRO}~\cite{wang2024mmlu} is an enhanced benchmark based on MMLU~\cite{hendrycks2020measuring}, to evaluate the language understanding abilities across broader and more challenging tasks.
(7) \textbf{MixEval}~\cite{ni2024mixeval} collects user queries from the web and matches them with similar queries from existing benchmarks, to bridge the gap between real-world user queries and ground-truth-based evaluation.
(8) \textbf{OpenLLM}~\cite{open-llm-leaderboard-v2} consists of commonly used datasets such as IFEval, BBH, MATH, GPQA, and MUSR, which compares LLMs in their own open and reproducible settings.

$\bullet$ \emph{Single-LLM Judge based Benchmarks.}

(1)\textbf{BiGGen Bench}~\cite{kim2024biggen} evaluates 9 core capabilities of LLMs, including instruction following, planning, reasoning, and others, using GPT-4 as the judge model along with instance-specific evaluation criteria. Meanwhile, (2)~\textbf{BiGGen Bench (Prometheus 2)} employs Prometheus 2~\cite{kim2024prometheus} as the judge model, serving as a complement to the original benchmark.
(3) \textbf{Alpaca Eval 2.0}~\cite{dubois2024length} employs GPT-4-Turbo as the judge and computes the win rates of the LLMs against GPT-4-Turbo for ranking.
(4) \textbf{WildBench}~\cite{lin2024wildbench} compares LLMs with three baseline models: GPT-4-Turbo, Claude3-Haiku, and Llama-2-70B on 1024 challenging real-world tasks. GPT-4-turbo is used as a judge to evaluate all the LLM pairs.

$\bullet$ \emph{Multi-LLM Judge based Benchmarks.}

(1) \textbf{PRD}~\cite{li2023prd} uses peer LLMs for weighted rankings of all LLMs, enabling fairer and more accurate assessments.
(2) \textbf{Auto Arena}~\cite{zhao2024auto} employs a committee of five strongest LLMs to evaluate other LLMs across 8 task categories.

\section{Evaluation Settings}\label{sec:evaluation_settings}
To provide a comprehensive comparison, we design three settings that evaluate 15, 30, and 66 LLMs, respectively, and report the performance on the overall and math dimensions.
Following existing work~\cite{ni2024mixeval}, we compute the Spearman Correlation between the ranking list from all benchmarks and the latest Chatbot Arena leaderboard.
Since Chatbot Arena rankings are based on human annotation, this approach allows us to estimate the correlation between automatic evaluation and human preferences.
Considering that the set of evaluated LLMs differs among benchmarks, we select the shared set of LLMs between the benchmark and Chatbot Arena to compute the correlation. 

\section{Implementation Details}\label{sec:imp_details}
For PRD, we re-implement it using the same hyper-parameter setting in the original paper. For other baseline methods, we collect the results from their official leaderboards.
For our De-Arena, we construct nine fine-grained dimensions using the data selection method in Section~\ref{sec:question_selection}, namely math algebra, math geometry, math probability, logic reasoning, social reasoning, science chemistry, science biology, science physics, and MT-bench.
We involve 15 open-source models in the data selection process to reduce the time cost.
For evaluation, we test 66 models in total and set the window size for fine-grained reranking to 1.
In the main experiments, we use the average rank of all nine dimensions as our final Overall rank, and the average rank of the three math sub-dimensions as our final Math rank. 
In the evaluation stage, for each benchmark, we identify and select the most relevant dimension provided by that benchmark and compare its results with the Chatbot Arena's Overall and Math dimensions, for calculating the correlation.
\section{De-Arena Leaderboard}
We show the detailed ranking results (\ie Elo scores) of all the evaluation dimensions from our De-Arena leaderboard in Table~\ref{tab:bigtable}.
It consists of the results from the dimensions of MT-Bench, Math (including Algebra, Probability, and Geometry three sub-dimensions), Reasoning (including Social and Logic), Science (including Biology, Chemistry, and Physics).
We also show the average scores for all dimensions as the overall score.
With the fine-grained sub-dimensions about math, reasoning, and science, we can have a comprehensive understanding of the detailed capabilities of LLMs, enabling to select the most suitable ones in specific tasks and scenarios.


\section{Visualization of Win-rate Distribution}
\begin{figure*}
    \centering
    \includegraphics[width=1\linewidth]{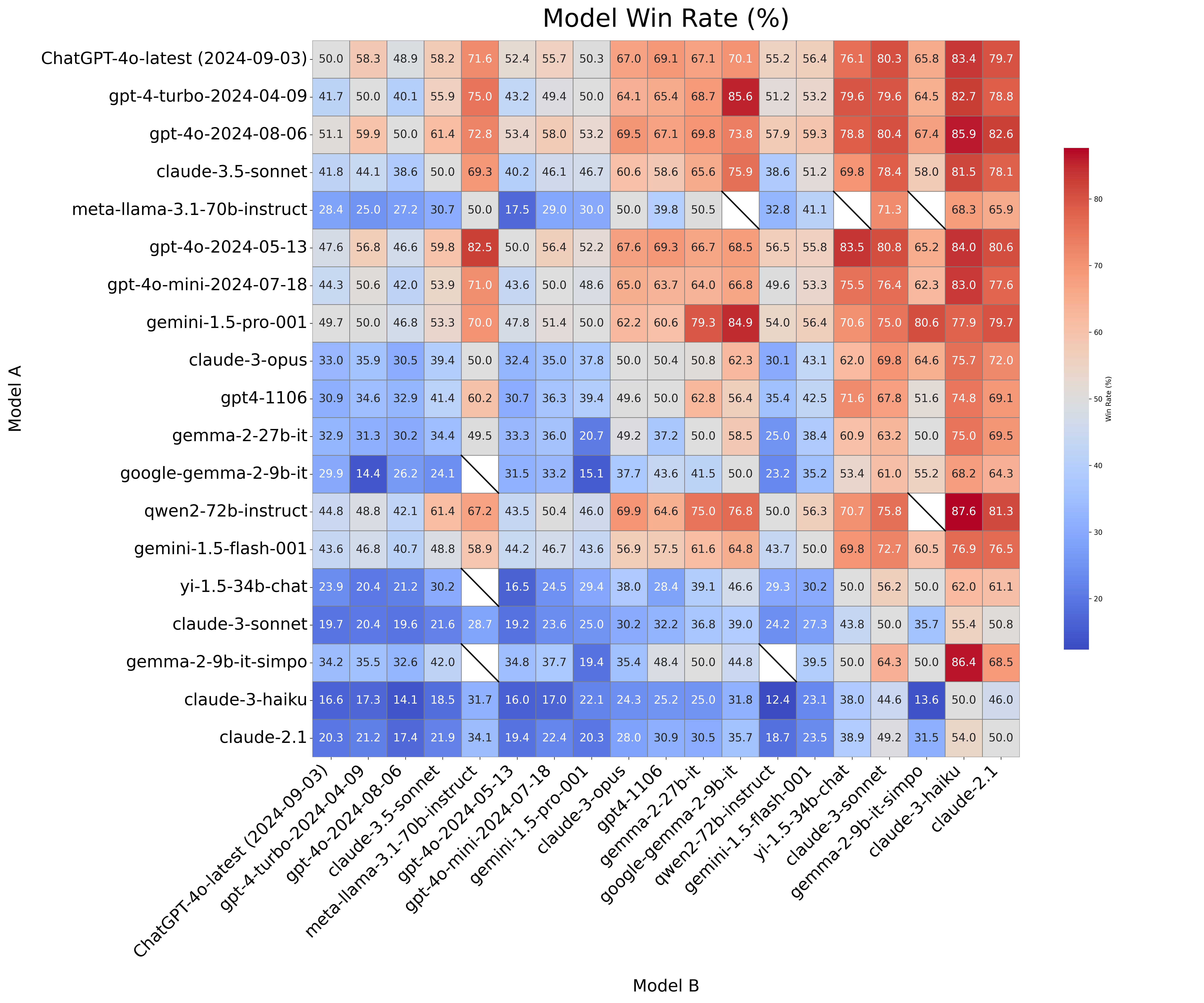}
    \caption{Win-rate distribution map for the evaluated LLMs using De-Arena.}
    \label{fig:winrate_map}
\end{figure*}

To better understand the results of our De-Arena, we also collect the win-rate of all LLM pairs and draw the distribution map in Figure~\ref{fig:winrate_map}.
We can see that the neighboring models in the ranking list (close to the diagonal), generally have near 50\% win rates.
It indicates that they are the more hard-to-distinguish ones than others with long distances, and they need more times of comparisons for determining their position in the ranking list.
This echos to the results in Figure~\ref{fig:compare_count_map}, where we can see that these neighboring models are also assigned with more comparison counts in our approach.

\begin{table*}[ht]
\caption{De-Arena Leaderboard on nine fine-grained dimensions.}
\centering
\large
\resizebox{\textwidth}{!}{%
\begin{tabular}{lcccccccccc}
\hline
\textbf{Model} & \textbf{Avg} & \textbf{Algebra} & \textbf{Probability} & \textbf{Geometry} & \textbf{Social} & \textbf{Logical} & \textbf{Biology} & \textbf{Chemistry} & \textbf{Physics} & \textbf{MT-Bench} \\
\hline
ChatGPT-4o-latest (2024-09-03) & 93.97611237 & 93.25223834 & 90.8076583 & 85.98448817 & 86.71827446 & 100 & 100 & 100 & 98.62585524 & 90.39649681 \\
o1-mini & 93.49965051 & 100 & 100 & 100 & 89.87352647 & 94.97639081 & 95.62385008 & 75.21644169 & - & 92.30699507 \\
o1-preview & 91.78818444 & 93.72198245 & 94.02361667 & 89.49116499 & 89.48962613 & 93.65959665 & 94.74772613 & - & - & 87.38357803 \\
yi-lightning & 90.39547454 & 92.29959103 & 93.09112725 & 86.21341092 & 92.57263596 & 88.97256711 & 88.06702943 & 91.27789372 & 88.37342652 & 92.69158893 \\
glm-4-plus & 89.83614921 & 89.16383214 & 81.98647507 & 83.72554158 & 100 & 94.93241923 & 90.77110696 & 88.62203483 & 93.85598831 & 85.46794477 \\
gpt-4o-2024-05-13 & 87.24525238 & 93.62469055 & 88.55622014 & 86.39564946 & 77.75644714 & 85.5560242 & 88.41283385 & 87.05454646 & 96.41218435 & 81.43867531 \\
claude-3.5-sonnet-20241022 & 86.18487078 & 85.70322602 & 85.82789712 & 81.08853546 & 89.24744502 & - & 83.01772535 & 89.50095243 & 100 & 75.09318483 \\
claude-3.5-sonnet & 86.0533705 & 84.97424618 & 80.51555832 & 81.47663229 & 92.88930807 & 88.7379798 & 85.15229796 & 87.25551817 & 91.66450066 & 81.81429306 \\
nemotron-70b & 85.70411922 & 82.07750032 & 81.70841233 & 76.85175181 & 94.53145012 & 88.98770368 & 85.27553532 & 76.19278843 & 85.71193093 & 100 \\
gpt-4-turbo-2024-04-09 & 85.44134591 & 88.24008459 & 87.79500275 & 83.43704168 & 82.55003583 & 86.58929024 & 87.99407319 & 79.24268595 & 90.17692838 & 82.94697059 \\
gemini-1.5-pro-001 & 83.84274456 & 93.28632947 & 79.80140526 & 85.73710469 & 85.13995771 & 79.22139256 & 86.46757085 & - & 87.24190789 & 73.84628802 \\
gpt-4o-2024-08-06 & 83.58357517 & 95.45825213 & 86.52136129 & 89.90076493 & 74.61636352 & 77.39296389 & 77.05217011 & 80.41289146 & 92.27434298 & 78.62306627 \\
meta-llama-3.3-70b-instruct & 81.64213667 & 88.05191213 & 81.50070219 & 76.01034051 & 93.37347192 & 80.58107374 & 80.49115306 & 72.91567181 & 82.81423433 & 79.04067033 \\
gpt-4o-mini-2024-07-18 & 81.62911065 & 91.90507855 & 87.46593196 & 83.73769897 & 75.21167553 & 78.08781194 & 75.84538799 & 76.10327227 & 89.37924614 & 76.92589247 \\
llama-3.1-tulu-3-70b & 80.69260189 & 83.20202637 & 80.99912888 & 81.36555176 & 73.0850881 & 81.7890594 & 83.98513871 & 79.25643379 & 82.99883141 & 79.55215858 \\
gemini-1.5-flash-001 & 78.66833532 & 83.971693 & 75.58450378 & 81.10124024 & 82.10090965 & 75.78605523 & 78.77971787 & 79.59254762 & 79.80184306 & 71.29650746 \\
qwen2-72b-instruct & 78.48455626 & 95.56883255 & 86.58621783 & 83.69563472 & 69.74028595 & 73.75123543 & 73.12077365 & 73.36129808 & 78.81838628 & 71.71834187 \\
claude-3-opus & 78.10866978 & 77.77044272 & 77.4873327 & 74.31586807 & 82.3069533 & 79.40132155 & 75.88174054 & 80.34596587 & 82.54044867 & 72.9279546 \\
gemma-2-9b-it-simpo & 77.31789719 & 77.09186481 & 69.27464096 & 72.17173541 & 83.46348374 & 68.03070404 & 86.46613755 & 81.56961063 & 80.47500035 & - \\
gpt4-1106 & 77.14331181 & 82.62125793 & 76.45691542 & 75.25907271 & 67.72358159 & 76.19284396 & 78.53497849 & 73.3599233 & 82.95606466 & 81.18516826 \\
gemma-2-27b-it & 74.9744008 & 79.32429033 & 74.68913171 & 72.95412253 & 75.46769179 & 75.72526971 & 77.38686738 & 74.41991612 & 74.4943143 & 70.30800333 \\
meta-llama-3.1-70b-instruct & 74.81908103 & 82.65410121 & 75.1056567 & 73.267449 & 71.7164733 & 75.28740754 & 72.90888115 & 71.61528424 & 75.99739508 & - \\
yi-1.5-34b-chat & 71.26419157 & 67.61590435 & 71.75031051 & 68.92660537 & 75.57098544 & 69.26273322 & 77.29432239 & 66.96425091 & 70.395733 & 73.59687895 \\
google-gemma-2-9b-it & 70.77950078 & 69.24880738 & 73.41916217 & 70.79663106 & 80.30676405 & 76.08614138 & 74.7058259 & 63.4394147 & 60.1404845 & 68.87227591 \\
llama-3-70b-instruct & 70.20905093 & 63.68177551 & 63.49465684 & 62.46945654 & 72.63114143 & 77.9502143 & 74.4856376 & 74.30515962 & 75.03715503 & 67.82626154 \\
claude-3-sonnet & 69.41240667 & 62.36474989 & 62.50401897 & 61.45200594 & 75.6759593 & 70.94830758 & 77.32499969 & 71.37876596 & 71.92772412 & 71.13512857 \\
claude-3-haiku & 65.60203719 & 59.59236589 & 58.39414152 & 56.83290624 & 76.22751044 & 66.38556042 & 73.9756324 & 65.90048762 & 66.54094904 & 66.56878114 \\
llama-3.1-tulu-3-8b & 64.13924992 & 74.7934719 & 69.57294421 & 72.08700434 & 49.9473266 & 42.15176782 & 68.92413911 & 62.85180162 & 70.70292737 & 66.2218663 \\
qwen1.5-72b-chat & 62.17737731 & 71.70038733 & 60.5941344 & 65.90803703 & 65.20990412 & 38.87873049 & 63.5728462 & 63.18615677 & 62.19290741 & 68.35329203 \\
claude-2.0 & 60.07907816 & 53.95876732 & 54.59652028 & 57.90072173 & 69.89646107 & 58.37909426 & 65.6804687 & 60.83364251 & 66.59410047 & 52.87192713 \\
ministral-8b-it & 59.8488688 & 61.63242352 & 58.725246 & 69.64691693 & 52.54413325 & 59.03562429 & 62.66986324 & 57.89155735 & 55.91787053 & 60.57618413 \\
qwen1.5-32b-chat & 59.06682296 & 70.78787721 & 60.35928986 & 61.69634576 & 57.70118438 & 41.23076181 & 59.40325534 & 54.81117145 & 63.07920208 & 62.53231871 \\
meta-llama-3.1-8b-instruct & 58.47833266 & 65.33759004 & 60.96410289 & 63.03060773 & 48.09357333 & 56.38560314 & 59.60200084 & 55.86477681 & 50.78809065 & 66.23864855 \\
claude-2.1 & 58.08029878 & 58.04833618 & 51.47446438 & 61.04594773 & 46.53270297 & 61.44324545 & 63.62055392 & 56.78385236 & 68.43538787 & 55.33819815 \\
qwq-32b-preview & 57.7441793 & 71.64275748 & 65.35384041 & 58.87771141 & 48.61002573 & 54.23148211 & 53.47734778 & 46.74960749 & 56.92639997 & 63.8284413 \\
qwen2.5-1.5b & 53.92178189 & 80.25022034 & 71.83262853 & 72.94135308 & 43.70205398 & 36.95513268 & 44.12989317 & 43.1067093 & 52.16952909 & 40.2085168 \\
llama3-8b-instruct & 52.91726564 & 46.66264507 & 40.01343217 & 44.11394158 & 51.5703736 & 58.94572886 & 60.93646636 & 57.92241191 & 56.85879525 & 59.23159592 \\
starling-lm-7b-beta & 52.3307152 & 52.84886768 & 46.5522985 & 51.13239206 & 51.61448852 & 48.63101151 & 58.5348304 & 50.50725399 & 44.75096252 & 66.40433164 \\
qwen1.5-14b-chat & 51.7844422 & 60.51769951 & 48.74784967 & 53.57316657 & 46.29169097 & 35.3542195 & 52.46656265 & 50.86736939 & 54.73141043 & 63.51001115 \\
mistral-8x7b-instruct-v0.1 & 51.55563155 & 51.29291916 & 48.27727168 & 49.19273323 & 42.54964874 & 39.00128601 & 62.82428402 & 56.26412563 & 55.53066816 & 59.06774734 \\
gpt3.5-turbo-0125 & 50.18821862 & 65.17729038 & 58.53802397 & 62.28075511 & 41.45152348 & 25.14406855 & 46.55809712 & 51.41965985 & 50.59440643 & 50.53014266 \\
command-r-(08-2024) & 48.62671709 & 46.83436386 & 42.80091277 & 52.40409927 & 46.35774401 & 27.95956428 & 55.60311141 & 51.79248215 & 50.9690958 & 62.9190803 \\
openchat-3.5-0106 & 47.43100598 & 50.5316225 & 46.18623367 & 46.49040206 & 43.70677816 & 39.44367439 & 55.48276235 & 43.37252958 & 43.15194557 & 58.51310554 \\
gemma-2-2b-it & 46.87564068 & 38.8089853 & 39.96987681 & 45.64812266 & 60.09115358 & 41.30384584 & 62.65299225 & 38.62991027 & 24.78287364 & 69.99300579 \\
command-r-(04-2024) & 45.65347002 & 33.00394744 & 37.6909176 & 38.12742778 & 46.5288232 & 36.53973546 & 56.21337898 & 52.08000978 & 52.33467243 & 58.36231751 \\
gemini-1.0-pro-001 & 45.55337569 & 51.6275763 & 37.0959279 & 50.19526563 & 29.97602848 & 27.69158385 & 43.65118076 & 55.27507409 & 60.00746246 & 54.46028173 \\
openchat-3.5 & 45.32554772 & 46.43700574 & 40.73751634 & 46.11460291 & 48.03856938 & 37.81208901 & 43.51031422 & 44.62699512 & 44.5992235 & 56.05361328 \\
gemma-1.1-7b-it & 45.08073221 & 45.73849545 & 38.11118608 & 39.98693207 & 30.0216809 & 38.86595628 & 52.31782605 & 52.65411393 & 51.64818503 & 56.38221413 \\
starling-lm-7b-alpha & 42.04507999 & 42.84399348 & 40.46448993 & 39.9951744 & 45.99949677 & 32.59182068 & 46.24425706 & 38.73575836 & 35.37621299 & 56.15451622 \\
mistral-7b-instruct-2 & 36.96477144 & 26.08274886 & 32.40740394 & 24.48546791 & 33.65114071 & 30.10038709 & 46.41225983 & 40.51553273 & 45.39716673 & 53.63083522 \\
llama-3.2-3b-it & 35.00733269 & 58.69046044 & 44.10665702 & - & 16.95381455 & 12.75038565 & 28.53090391 & 37.646318 & 33.86613149 & 47.51399047 \\
vicuna-33b & 34.88330747 & 27.47528016 & 29.6760891 & 25.60995953 & 42.34355432 & 23.15924552 & 47.12097365 & 33.63396701 & 34.13729337 & 50.79340454 \\
gemma-7b-it & 32.62523838 & 24.22405867 & 24.14389506 & 28.60141294 & 36.71945984 & 27.42936244 & 34.3566972 & 35.61243194 & 35.37364385 & 47.16618346 \\
llama-3.2-1b-it & 31.50652515 & 48.19647744 & 35.35718488 & 38.4098706 & 22.37884882 & 14.82696436 & 30.66596363 & 28.17342475 & 18.52660051 & 47.02339137 \\
smollm2-1.7b & 29.77883794 & 41.41094808 & 34.66233542 & 40.86703593 & 24.65541733 & 14.47158568 & 28.68894368 & 21.27370007 & 24.5651071 & 37.41446812 \\
mistral-7b-instruct-1 & 25.18022546 & 24.28958881 & 30.91573756 & 21.30326088 & 25.0166404 & 19.58058203 & 27.73220177 & 21.1732995 & 19.77456982 & 36.83614841 \\
vicuna-13b & 24.54431725 & 23.01589211 & 17.95905259 & 22.98158082 & 24.22989261 & 19.67247727 & 34.52179233 & 23.25790538 & 23.32247878 & 31.93778333 \\
gemma-1.1-2b-it & 22.23412728 & 13.45837587 & 7.880815666 & 21.50934176 & 28.61649971 & 11.99679124 & 28.48075179 & 22.20401391 & 24.68483409 & 41.27572145 \\
qwen1.5-4b-chat & 21.64067671 & 33.87561687 & 19.1517295 & 25.22230824 & 32.39522118 & 14.53055119 & 14.20140305 & 18.12674834 & 17.35368952 & 19.90882249 \\
llama2-7b-chat & 20.37432805 & 12.45837033 & 8.124426409 & 9.546107548 & 23.34687358 & 19.83603456 & 30.18764134 & 22.86497137 & 15.31628193 & 41.68824533 \\
llama2-13b-chat & 19.7016256 & 7.162722477 & 13.89583417 & 9.340440711 & 23.72229041 & 24.18676702 & 26.80803286 & 18.33849082 & 12.95980209 & 40.90024986 \\
gemma-2b-it & 18.70510718 & 12.73154473 & 7.739314245 & 25.55963926 & 20.20062881 & 6.432548745 & 19.03185637 & 18.14493366 & 20.88252343 & 37.62297534 \\
vicuna-7b & 17.98491319 & 8.77942567 & 8.020707561 & 9.731584621 & 22.36270901 & 14.08926915 & 28.51353075 & 21.44761184 & 16.70282166 & 32.21655842 \\
zephyr-7b-beta & 14.57067856 & 16.3116851 & 10.09469635 & 13.2538558 & 0 & 9.979969635 & 22.337812 & 20.35092805 & 18.4166841 & 20.39047603 \\
koala-13b & 9.409161591 & 2.531969921 & 4.683320295 & 2.659127343 & 20.54769134 & 9.810835924 & 19.26051579 & 5.954307371 & 3.708961286 & 15.52572505 \\
openassistant-pythia-12b & 0.405780038 & 0 & 0 & 0 & 3.652020346 & 0 & 0 & 0 & 0 & 0 \\
\hline
\end{tabular}
}
\label{tab:bigtable}
\end{table*}

\end{document}